\documentclass{article}

\usepackage{arxiv}

\usepackage[utf8]{inputenc} 
\usepackage[T1]{fontenc}    
\usepackage{hyperref}       
\usepackage{url}            
\usepackage{booktabs}       
\usepackage{amsfonts}       
\usepackage{nicefrac}       
\usepackage{microtype}      
\usepackage{lipsum}		
\usepackage{graphicx}
\usepackage{natbib}
\usepackage{doi}

\usepackage[utf8]{inputenc} 
\usepackage[T1]{fontenc}    
\usepackage{hyperref}       
\usepackage{url}            
\usepackage{booktabs}       
\usepackage{amsfonts}       
\usepackage{nicefrac}       
\usepackage{microtype}      

\title{Site-specific weed management in corn using UAS imagery analysis and computer vision techniques}

\author{ \href{https://orcid.org/0000-0002-5417-6744}{\includegraphics[scale=0.06]{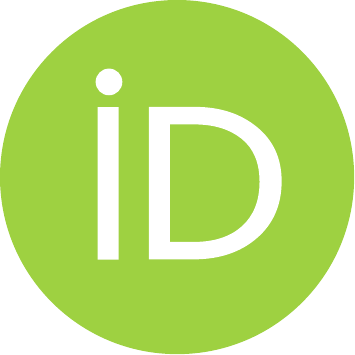}\hspace{1mm}Ranjan Sapkota}\thanks{} \\
	Center for Precision \& Automated Agricultural Systems\\
	Washington State University\\
	24106 N. Bunn Rd, Prosser, 99350, Washington, USA \\
	\texttt{ranjan.sapkota@wsu.edu} \\
	\And
	\href{https://orcid.org/0000-0000-0000-0000}{\includegraphics[scale=0.06]{orcid.pdf}\hspace{1mm}John Stenger} \\
	Agricultural and Biosystems Engineering\\
	North Dakota State University\\
	1221 Albrecht Blvd, Fargo, 58102, North Dakota , USA \\
 \And
	\href{https://orcid.org/0000-0000-0000-0000}{\includegraphics[scale=0.06]{orcid.pdf}\hspace{1mm}Michael Ostlie} \\
	NDSU Carrington Research Extension Center\\
	North Dakota State University\\
	Carrington, ND 58421-0219, USA \\
\And
	\href{https://orcid.org/0000-0000-0000-0000}{\includegraphics[scale=0.06]{orcid.pdf}\hspace{1mm}Paulo Flores} \\
	Agricultural and Biosystems Engineering\\
	North Dakota State University\\
	1221 Albrecht Blvd, Fargo, 58102, North Dakota , USA \\
}

\renewcommand{\shorttitle}{\textit{arXiv} Template}

\begin{document}
\maketitle

\begin{abstract}
Currently, applying uniform distribution of chemical herbicide through a sprayer without considering the spatial distribution information of crops and weeds is the most common method of controlling weeds in commercial agricultural production system. This kind of weed management practice lead to excessive amounts of chemical herbicides being applied in a given field. The objective of this study was to perform site-specific weed control (SSWC) in a corn field by: 1) using a unmanned aerial system (UAS) to map the spatial distribution information of weeds in the field; 2) creating a prescription map based on the weed distribution map, and 3) spraying the field using the prescription map and a commercial size sprayer. In this study, we assumed that plants growing outside the corn rows are weeds and they need to be controlled. The first step in implementing such an approach is identifying the corn rows. For that, we are proposing a Crop Row Identification (CRI) algorithm, a computer vision algorithm that identifies corn rows on UAS imagery. After being identified, the corn rows were then removed from the imagery and remaining vegetation fraction was classified as weeds. Based on that information, a grid-based weed prescription map was created and the weed control application was implemented through a commercial-size sprayer. The decision of spraying herbicides on a particular grid was based on the presence of weeds in that grid cell. All the grids that contained at least one weed were sprayed, while the grids free of weeds were not. Using our SSWC approach, we were able to save 26.23\% of the land (1.97 acres) from being sprayed with chemical herbicide compared to the existing method. This study presents a full workflow from UAS image collection to field weed control implementation using a commercial size sprayer, and it shows that some level of savings can potentially be obtained even in a situation with high weed infestation, which might provide an opportunity to reduce chemical usage in corn production systems. 
\end{abstract}

\keywords{Remote Sensing \and Unmanned Aerial Vehicle \and Weed management \and site-specific weed management \and computer vision \and Drone \and GIS \and precision agriculture \and weed control \and UAV \and UAS \and sensors }

\section{Introduction}
About 1 billion pounds of conventional pesticides are used each year to control undesirable vegetation such as weeds in the agricultural production system of the United States (U.S.), which amount to a cost of almost \$9 billion \cite{USGS1,sow-2022}. One of the major reasons for excessive use of chemicals in agriculture is the conventional method of spraying herbicides, often called a blanket application for weed control, which accounts for 59\% of the major pesticide expenditures in economic crops in North America \cite{sharma2019worldwide}. The blanket application implies spraying herbicide uniformly across the field without considering the spatial distribution of the weeds, which often causes overuse of chemicals.

\par
In addition to being harmful to the environment \cite{fugere2020community, vieira2020herbicide}, chemical herbicides can be harmful to human health as they have been linked to cancer\cite{koutros2013risk}, attention deficit hyperactivity disorder (ADHD)\cite{polanska2013review}, alzheimer’s\cite{yan2016pesticide}, DNA damage, cardiovascular diseases, neurological disorders, and reproductive disorders\cite{kaur2018occupational}. Moreover, excessive chemical use in agriculture has the potential to harm nervous system, endocrine system, and reproductive system of human\cite{unknown-author-2011}.  On the other hand, excessive use of chemicals in agriculture have become one of the major reason for surface and groundwater pollution \cite{srivastav2020chemical,zhang2018impact,kumar2019chapter,hasan2019water}, resulting big threat to the ecosystem and public health. Although organic farming is seen by many as a potential solution for the excessive use of chemicals in agriculture, it is not a silver bullet to meet the United Nations sustainable development goals by 2030 \cite{eyhorn2019sustainability}. Therefore, to maintain environmental sustainability and human health, agricultural production systems are in dire need of technologies that can significantly reduce the amount of chemicals currently being used, without compromising crop yields.
\par
An alternative way to neglect chemical use for weed control in agricultural production system is through physically removing weeds by using robotics platforms.  Some of the latest automated robots such as the autonomous weed robot by “ecoRobotix” (ecoRobotix, Vaud, Switzerland) and “Deepfield Robotics” (Deepfield Robotics, Renningen, Germany), use ground-based machine vision and image processing techniques to perform site-specific weed control (SSWC) in row crops \cite{bib14}. However, these robotic solutions are not effective to be deployed in a large commercial agricultural field (hundreds or thousands of acres).
\par 
To overcome that issue, some companies have developed and commercialized solutions such as “Weedseeker®” (Trimble agriculture, California, USA), and “WEED-it” (WEED-IT, Steenderen, Netherlands), which use optoelectronic sensors to measure the reflection intensity of vegetation, allowing them to discriminate vegetation from soil background  \cite{peteinatos2014potential}. The shortfall of those products, at the time of this writing, is that they cannot discern weeds from crops early in growing to allow such technologies to be used for weed control. Recently, John Deere (Illinois, USA) launched See \& Spray Ultimate®, a sprayer that enables detection and control of weeds during the growing season in corn, soybean, and cotton. More details regarding the field performance of that technology remain to be seen since the sprayers fitted with the system will be available only in 2023\cite{john-deere}.
\par
Another approach of performing SSWC is possible through the use of remote sensing technology to accurately map the weed distribution information across a field and integrate that with a spraying platform \cite{bib13}. The main idea here is to spray herbicide only to those areas where weeds are present \cite{jin2021weed}. To accomplish that, one would need a high-resolution imagery to discern plants growing in the field as weeds and crop. Since the satellite imagery would present some challenges to discern weeds from crops, due to its limitations in providing adequate spatial resolution\cite{herwitz2004imaging}, the use of UAS with high-resolution cameras seems to be a promising platform, which can be processed and analyzed to obtain an accurate weed distribution map from a field \cite{bib18}. For that approach to work, is imperative to detect weeds accurately early in the growing season. In most crop weed management programs, weed treatment is recommended at the early growth stages of both crops and weeds. Mapping weeds at that stage can be a challenging task because of four main reasons: 1) weeds are not uniformly distributed across the field, which necessitates working at a single pixel size on the image \cite{mani2021remote}, 2) crop and weeds have the same or similar reflectance properties, 3) interference of soil background \cite{thorp2004review}, and 4) very small size weeds which are not captured by the sensor’s insufficient camera \cite{lopez2016early}. 
\par

One of the earliest studies related to SSWC using UAS \cite{torres2013configuration} reported about the possibility of the use of UAS imagery and imagery analysis techniques to perform accurate discrimination between weeds and sunflower at early growth stages, which is key to implement any SSWC. More recently, the use of UAS in agriculture for SSWC have become a topic of several publications in specialized literature related to agriculture \cite{mink2018multi, pena2013weed, gao2018fusion, sapkota2020mapping, rasmussen2019pre, louargant2018unsupervised, zisi2018incorporating, rasmussen2020novel, jurado2019papaver}. Around the same time, more sophisticated and powerful image analysis algorithms, such as machine learning (ML) and deep learning (DL) have being developed and adapted for agricultural applications, and they have gained space as effective ways to process large amounts of data in agriculture \cite{koul2021machine,liakos2018machine}. 
\par
Identification of crop rows in an UAS imagery using ML and DL have become another subject for several recent studies \cite{pang2020improved,varela2018early,ronchetti2020crop, de2018automatic, torres2021early, vong2021early, bah2019crownet, osco2021cnn, basso2020uav, wang2021convolutional}. Although, most of these studies have been able to identify crop rows, and discriminate between weeds and crops during the early stage, however, these all studies are limited to virtual simulations only. Most of these reports regarding the use of UAS imagery for SSWC fall short of providing a real word solution for SSWC, since most of them just report on the capability of different algorithms to separate weeds from crops. Moreover, most of these studies regarding crop row detection, were able to identify the rows as lines with great accuracy \cite{ronchetti2020crop, de2018automatic, torres2021early, vong2021early, bah2019crownet, osco2021cnn, basso2020uav, wang2021convolutional}, however, the authors could not provide any insights on how to use those detected crop rows in practical agriculture. Additionally, most of the reports are performed in a small area of agricultural land which would probably raise questions about the research validity in a bigger agricultural area. 
\par
Among the row crops grown the United States (US), corn is the primary, largest and most important feed grains, accounting for more than 95\% of total feed grain production in the  \cite{bib1}. Corn is a major crop for the state of North Dakota (ND), US. In 2020, ND farmers grew corn in 1.5 million hectares of corn in ND \cite{bib4}. Weed competition is a major cause of corn yield losses in ND, and chemical herbicide is the most widely used option for weed control \cite{bib5}. 
\par
In order to optimize weed control efficacy, the use of pre- and post-emergence chemical herbicide treatments have been proven to be the most effective weed control strategy \cite{bib6, bib7}. At present, most of the farmers apply a blanket application of herbicides for the post-emergence weed control in corn, with no consideration regarding the spatial distribution of weeds across the field. That application often results in an overuse of chemical herbicides. The cost of post-emergence herbicide application (product + application) in ND is usually around \$8-12/ha (2020 growing season). If one could use UAS imagery to generate a weed distribution map, to implement SSWC, and apply chemicals to only those areas where weeds are present in the field, that could lead to significant savings on chemicals for corn growers. Using the area grown with corn in 2020 in ND and the cost associated with chemical application, those saving could reach as much as \$5.5-7.5 million, with as little as a 10\% reduction in applied acreage. In addition to the financial savings, there are environmental savings, which are harder to put a number to it, that can be realized using applying less chemicals to the land.
\par
To our knowledge, this is the first study to report on the whole workflow, from data image collection to weed control in the field, to implement SSWC in corn by integrating UAS imagery with a commercial size sprayer. This study was carried out with two major objectives:
\begin{itemize}
\item to use UAS imagery to map and quantify post emergent weed infestation in a corn field
\item to integrate and evaluate that map as a prescription map for site-specific weed control using  a commercial size sprayer.
\end{itemize}

\section{Material and methods}
\begin{figure}[h]
    \centering
    \includegraphics[width=0.65\linewidth]{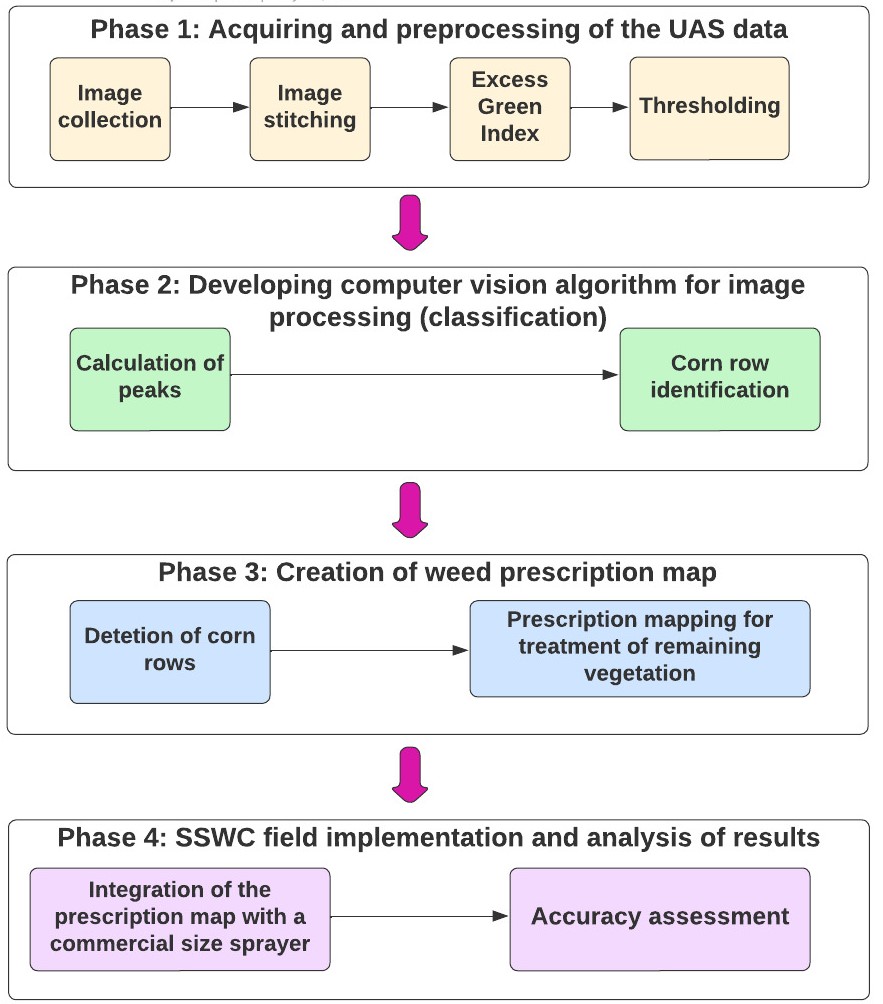}
    \caption{Workflow diagram followed to implement a site-specific weed control approach in a corn field}
    \label{fig:workflow}
\end{figure}

This study can be divided into four phases (Figure \ref{fig:workflow}); (1) acquiring and preprocessing UAS data, (2) develop a computer vision algorithm for corn row classification, (3) weed prescription map creation, and (4) field implementation of the prescription map.

\subsection*{Study site}\label{subsec9.1}
The study was carried out in a 17 hectares (approximate) corn field,  located at Carrington, North Dakota $\left(47.51^{\circ} N, 99.12^{\circ} W\right)$. The experimental land for this study was provided by North Dakota State University (NDSU) Carrington Research Extension Center (REC). According to the data provided by web soil survey (WSS), the soil in the experimental field was composed of heimdall and similar soil (42\%) , emrick and similar soil (37\%), and minor components 21\%. The field was planted with silage corn on May 12, 2021, with a 30-inch row spacing pattern. Prior to planting, on May 7, the field received a pre-emergence herbicide treatment (Verdict \cite{BASF}, 0.98 kg/ha), which was applied at a rate 140.3 L/ha. 

\subsection*{Data collection }\label{subsec9.2}
The UAS flights for image collection were carried out on June 14, 2021. A DJI Matrice 600 Pro (M600) (figure 2(a)(DJI, Shenzhen, China), outfitted with a Sony Alpha 7R II 42 Megapixel RGB camera (Sony City, Tokyo, Japan)  was used to capture the aerial images of the field as shown in figure 2(a). The camera has a 7952 × 5304 pixels (42.4 megapixels) sensor resolution, and a focal length of 35mm. Integration unit for the camera on the drone was made by FieldofView LLC (Fargo, North Dakota, USA), which manufactures a device  called GeoSnap PPK 2(b) that allows one to trigger the camera and geotag the images with PPK (postprocess kinematic) accuracy (2cm). Since distance between the research field and nearest CORS (Continuously Operating Reference Stations; Cooperstown, ND) was too far, which would cause a degradation of the geotag accuracy, an iG4 GNSS (Global navigation satellite system) base station (iGage Mapping Corporation, Salt Lake City, USA) (figure 2c) was used to implement the PPK correction to the geotags. The UAS was flown autonomously using flight missions created in Pix4DCapture app (IOS version) (PIX4D, Prilly, Switzerland). Flights were carried out at 350 feet above ground level (AGL), with the camera at nadir position, with 75\% overlap both front and side, and the UAS speed was adjusted (by the app) to allow 1.4 seconds interval between pictures. Altogether, three flights were carried to cover the experimental field, collecting a total of 2251 images.

All the data processing in this study was performed in a desktop computer that have Windows 10 Pro, version 20H2, a 64-bit operating system with 132 gigabytes of RAM. The desktop was provided with two switchable GPUs. One was Intel(R) UHD Graphics 630, and the other was NVIDIA GeForce RTX 2080 SUPER.
\begin{figure}[h]
    \centering
    \includegraphics[width=0.95\linewidth]{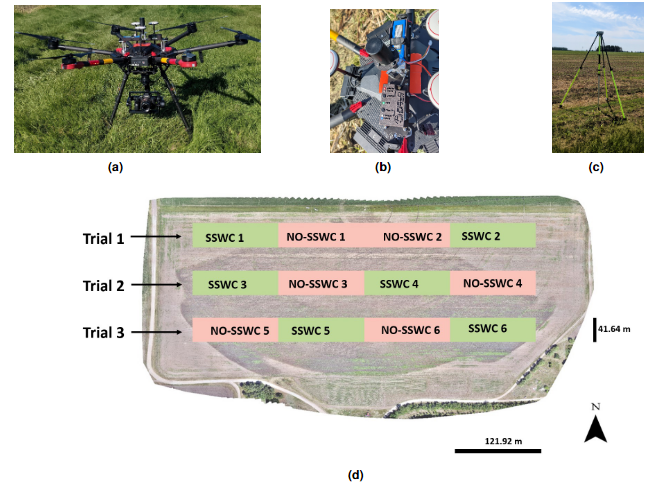}
    \caption{(a) DJI Matrice 600 Pro with a Sony Alpha 7R II, 42.4-megapixel RGB camera and Geosnap PPK system (top) mounted to it;   (b) detailed view of Geosnap PPK system mounted on top of the aircraft;   (c) ig4GNSS base station that was kept in ground for stationary position; and   (d) Orthomosaic generated over a corn field after the images captured by a Sony Alpha 7R II, flown at 350 ft AGL were stitched in Pix4D, and placement of the SSWC (green) and no-SSWC (pink salmon) treatment plots for the experimental design.}
     \label{fig:figure2}
\end{figure}

\subsection{Image preprocessing and stitching}
In order to generate accurate (2 cm accuracy for latitude and longitude, and twice that for elevation) geotag information for each image, the information collected by the GNSS base station  and GeoSnap device were processed using EzSurv software (Effigis, Montreal, Canada), which provided an output as “.csv” file with post-corrected geotag information for each picture. That file was then used during the images stitching process. Images were processed into an orthomosaic (figure 2 d) using image stitching software, Pix4DMapper (Pix4D, Prilly, Switzerland). It took 6 hours and 25 minutes for the software to generate the output as orthomosaic, which had an average ground sample distance (GSD) of 0.63 cm. That orthomosaic served as basis for all the subsequent analyses to implement our SSWC approach. Figure 3 shows the orthomosaic with the experimental units where both the SSWC and conventional (No-SSWC) treatments were applied.

\subsection{Field area experimental design}

A spatial field experiment area was designed on the orthomosaic to perform both the traditional method which farmers are currently using and our SSWC method. Altogether 3 trial plots of each 487.68m x 136.6m as shown in figure 2(d) were created. Each trial plot was then equally divided into four parts  (121.92m  x 41.64 m each), making 12 equal plots for the implementation of weed control methods. Among the 12 plots, 6 test plots were replications of our SSWC approach and the remaining 6 test plots were replications of the conventional method or no-SSWC approach. The spatial design for the experiment was created using a geographical information system (GIS) application, ArcGIS Pro (ESRI, California, USA). 
\par Once the two approaches of weed control was applied, the total sprayed area and non-sprayed area for the SSWC replications was accessed through the sprayer’s computer and compared with the no-SSWC replications. The performance of our method was then evaluated based on the land area where chemical herbicide was not sprayed as there was not presence of weeds. 

\subsection{Vegetation Identification}
In order to segment the vegetation fraction from the background on the orthomosaic, excess green index (ExGI) was calculated in ArcGIS Pro, using equation \ref{eq:1}.
\begin{equation}
ExGI = 2g-r-b 
 ;  \hspace{1cm}  where \hspace{0.4cm} (r=\frac{R}{R + G + B}
\hspace{10pt}
g =\frac{G}{R + G + B}
\hspace{10pt}
b =\frac{B}{R + G + B}
\label{eq:1})
\end{equation} 
where r, g, and b are the normalized values of the bands red, green, and blue respectively, and R, G, and B represent the pixel digital number values of the red, green, and blue color bands respectively. 
Once the ExGI was calculated, a threshold value (0.08) was manually/visually identified and applied to distinguish all green vegetation from the background. That resulted in a binary imagery with pixel intensity value 1 for the vegetation and 0 for the remaining background. The binary image was further processed to identify the corn rows.

\subsection{Algorithm development for crop row detection}
\begin{figure}[h!]
    \centering
    \includegraphics[width=0.9\linewidth]{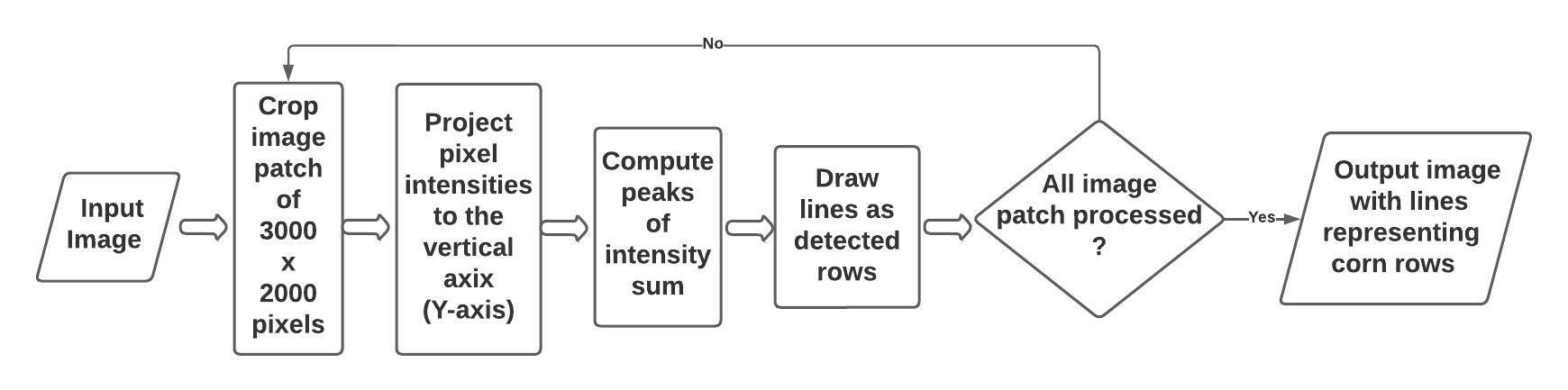}
    \caption{Flow diagram of Corn Row Identification (CRI) algorithm to detect lines over corn rows in binary imagery generated from UAS imagery.}
    \label{fig:4}
\end{figure}

We are proposing the use of a crop row identification (CRI) algorithm for the detection of corn rows in the binary orthomosaic imagery. In order to expedite the processing time, the CRI algorithm was applied only to the six SSWC treatment plots, which was around 3.05 hectares in area. Figure \ref{fig:4} shows the flow diagram for the CRI algorithm that is being proposed for corn row identification.

Pixel intensity along X-axis (columns) as shown in figure 4 (a) were summed and projected towards Y-axis (rows) in order to compute the local maximums (peaks) in Y-axis as shown in figure 4(b). Once the local maximums were computed for each row, straight lines perpendicular to Y-axis were drawn as shown in figure 4(c). The same process was repeated for all section that was inside the boundary of the SSWC treatment plots. 
\par Once the lines were drawn over the computed local maximums in each row, those lines were visually compared with the original corn rows (ground truth) on the imagery using ArcGIS Pro. Figure 4(d) shows the visual comparison of the generated outputs as true positive(TP), false positive(FP) and false negative(FN). That visual inspection was used to verify the performance of CRI algorithm for accurately identifying the corn rows from the UAS imagery.

\begin{figure}[h]
    \centering
    \includegraphics[width=0.95\linewidth]{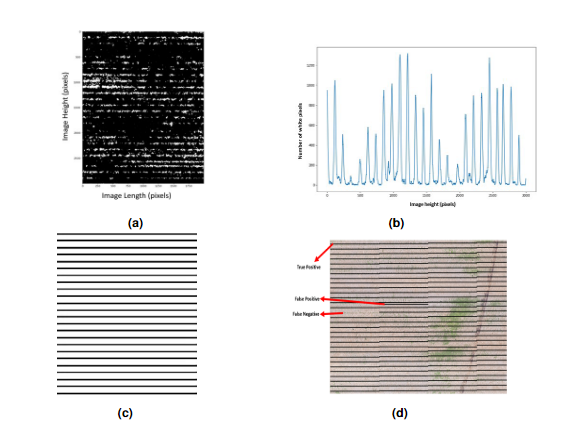}
    \caption{(a) A section of the binary ExG imagery (3000 X 2000 pixels) showing the horizontal orientation of rows,  b) local maximums (or peaks) for each corn row on the section of imagery, c) lines drawn over the each computed local maximum, and (d) Comparison of actual corn rows with the identified lines created by the CRI algorithm on a corn field imagery captured by Sony Alpha 7R II, flown at 350 ft AGL}
        \label{fig:figure4}
\end{figure}
\subsection{Weed mapping across SSWC treatment plots}
Our approach of mapping weeds is quite simple, and that is by design. All the vegetation that growing within a certain distance (buffer) from the identified rows are considered corn plants, and the remaining of the vegetation, growing mainly between the rows, are considered weeds. Hence, once the corn rows were identified, the next step was to identify the vegetation fraction that would be classified as weeds. Our approach was to identify and delete all corn plants from the treatment plots, and the remaining vegetation was then classified as weeds. In order to implement that approach, we created a 3.5 inches buffer on both sides of the corn row lines as shown in figure 5(a). This 3.5-inch buffer was wide enough to cover most of the corn plants, which were then deleted from the imagery, leaving behind the weeds present between the corn rows as shown in figure 5(b).
Since the weed pressure in the field was unusually high in the 2021 growing season, in order to implement the full workflow of the proposed approach for SSWC, from image collection to field spraying with a commercial size sprayer, the research team opted to overlay a grid of cells of 1.67 x 10 ft on the imagery. That resulted in an average value of 35\% of the cells free of weeds across the SSWC treatment. The decision of spray or no spray on a given cell was based on the presence of weeds in a given grid cell. Grid cells that contained weed were assigned a rate of 140.3 L/ha herbicide mix solution, while the cells free of weeds were assigned at a rate of 0 L/ha, which is no-spray. The idea is that sprayer would just shut-off the nozzles while travelling over the no-spray cells. Figure 5(c) shows the prescription shapefile created in ArcGIS Pro, which was converted to a prescription map in a later step. The red cells among the replications of SSWC treatment were free of weeds whereas the green cells contained weeds. 

\begin{figure}[h]
    \centering
    \includegraphics[width=0.95\linewidth]{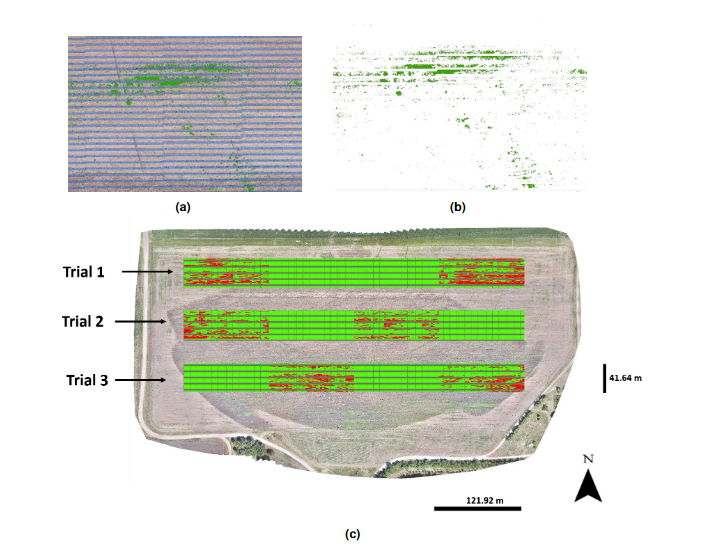}
     \caption{Buffered (3.5-inch on both sides) corn rows (a) that were removed from the imagery to create the weed map across the field (b), (c)Variable rate prescription map (green cells = 37.06 gal/ha and red cells= 0 gal/ha) generated in ArcGIS Pro based on UAS imagery collected over a corn field using a Sony Alpha 7R II camera, flown at 350 ft above ground level.}%
    \label{fig:figure5}%
\end{figure}

\subsection{Integration of Weed Map into a commercial sprayer}
A Case IH Patriot 4440 self-propelled sprayer (Racine, Wisconsin, USA), model year 2021, (figure 6a) was implemented to spray herbicide in this study. The sprayer was provided with an AIM command FLEX system, which was the latest technology from Case IH that enables consistent, flexible, and accurate spray application in commercial agriculture, regardless of speed and terrain. In addition, the system enhances control of liquid product flow and pressure more accurately than conventional rate controller and enables instant on/off of individual nozzles, with a nozzle valve diagnostic system.

\newpage The sprayer was set up with a Raven RX1 real time kinematic (RTK) receiver (Raven Technology, Sioux Falls USA), which operates at a frequency of 10Hz. Following Case IH engineer’s advice, the application speed for this study was kept below 11 km/hr  (10.5 km/hr actual speed application) to allow for the sprayer to maintain position accuracy resolution of at least one foot per 1/10th of a second or 10 ft per second. That cab computer requires a certain folder structure, so it can read prescription maps (Rx) from an external storage device, such as a thumb drive. AgSMS Advanced software (Ag Leader, Iowa, USA) was used to read the shapefile created in ArcGIS Pro and to convert that file into a prescription map (Rx map) following the format required by the Viper 4+ computer cab. Once the Rx map was brought into the Viper 4+ display, the regions indicated by green color was set to be sprayed with a 56.78 L/ha, which is the same rate that farmers would use for a conventional chemical application.
\begin{figure}[h]
    \centering
    \includegraphics[width=0.95\linewidth]{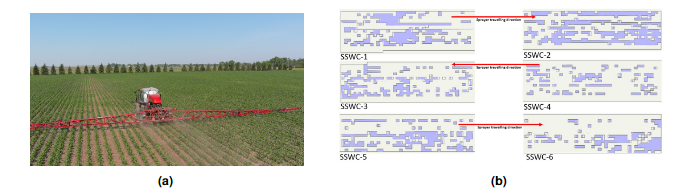}
    \caption{(a)Case IH Patriot 4440 series sprayer used to implement site-specific weed control in a corn field. The sprayer was equipped with an AIM command FLEX system, RTK GPS receiver, Viper 4+ cab computer, and 136.6 ft wide boom; (b) Overlap of Rx map, generated by AgSMS, where the regions enclosed by dark black lines where the no-spray regions (0 gal/ac), and the as-applied map. Purple color regions are the area where the sprayer turned its nozzles off while operating at 10.5 km/hr speed (approximate), while all the remaining area was sprayed (37.06 gal/ha)}%
    \label{fig:example1}%
\end{figure}

\subsection{Assessing the spraying performance in terms of spray accuracy and chemical savings}
Once the spraying operation was completed, an as-applied map was downloaded from the sprayer’s cab computer to compare with the prescription map in terms of spray and no-spray area. The spatial area recorded on the 6-SSWC treatment plots figure 6 (b) of the as-applied map where the sprayer did not spray any chemical herbicide was calculated and compared with the other 6 plots (no-SSWC) where conventional method was applied.

Additionally, application accuracy of the sprayer was evaluated based on it's ability to perform individual nozzle shut-off over the land where there was no presence of weeds. Based on the sprayed and not sprayed spatial area in the treatment zones between the prescription map and the as-applied map, the sprayer’s performance for chemical saving was evaluated. For that, the total area which was recorded as not-sprayed in the as-applied map was considered as Measured Value, and the total area which was set to be not-sprayed in the prescription map was considered as Expected Value. The accuracy of application in terms of spraying was calculated using equation 
 \ref{eq:sprayeraccuracy}: 
 \begin{equation}
Accuracy=\frac{measured\ value}{expected\ value}
\label{eq:sprayeraccuracy}
=\frac{area\ of\ not\ sprayed\ regions\ in\ as\ applied\ map}{area\ of\ no-spray\ regions\ in\ prescription\ map}
\end{equation}

\subsection{Post-harvest field analysis}
In order to investigate the experimental plots in terms of weed growth in a post harvest season, we collected a post harvest imagery on September 21, 2021, where image collection, image stitching, georeferencing, ExGI calculation, segmentation, and thresholding was carried out in a similar way as described for early-stage corn field data collection. The surface area of leftover and newly germinated weeds in the SSWC and conventionally approached test plots in the post harvest imagery was calculated and the datasets were analyzed using SAS PROC MIXED (SAS Institute, Cary, USA) with a mixed procedure using REML (restricted maximum likelihood) estimation. The area covered by weeds in the SSWC test plots was compared to the area of weeds present in six no-SSWC test plots using a pair-wise T-test with a significance level of 0.05 (p = 0.05).

\section{Results and discussion}
\subsection{CRI Algorithm performance for crop row detection}
Our simple computer vision algorithm "CRI (corn row identification) algorithm" was able to identify and draw corn rows from UAS imagery with great accuracy. There were 2313 true positive (TP), 12 false positive (FP), 0 true negative (TN), and 8 false negative (FN) values. Overall the CRI algorithm performed well when identifying corn rows based on a UAS imagery over 6.1 hectares land, with metrics: precision (0.9935), recall (0.9965), F1-score (0.9949), and accuracy (0.9901) all above 0.99. The algorithm processing time was 8 seconds using the desktop computer previously described. 
\par
Because of the very small size of some corn plants as shown in figure 7 (a), the algorithm failed to identify line over those regions. The row of corn plants that could not be identified in the figure is due to the effect of the threshold value (0.08) that was applied during image segmentation. Use of higher resolution camera that can map even the smallest size of corn plant could possibly solve this issue. Also, in very few regions, we have noticed two lines representing a single corn row as shown in figure 7 (b). This is because the corn rows are not perfectly straight inside the region of 3000 x 2000 pixels, specified in the CRI algorithm. Since the corn row is not straight over the part of image, two small peaks(local maxima) have been identified in this region and hence, the algorithm displayed 2 lines to represent a single corn row. Setting up the smaller region of interest during image processing could possibly solve this issue. 

\begin{figure}[h]
    \centering
    \includegraphics[width=0.95\linewidth]{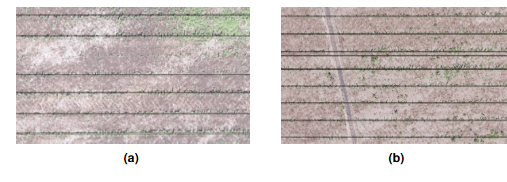}
   \caption{(a) Sample image where the proposed CRI algorithm failed to detect an existent corn row as a line on the ExGI imagery of a corn field, (b) Sample image where the proposed CRI algorithm detected two lines over a single corn row on the ExGI imagery of a corn field}%
    \label{figure7}%
\end{figure}
\newpage 
\subsection{Application accuracy and chemical savings using our SSWC method}
\begin{figure}[h]
    \centering
  \centering{{\includegraphics[width=13cm]{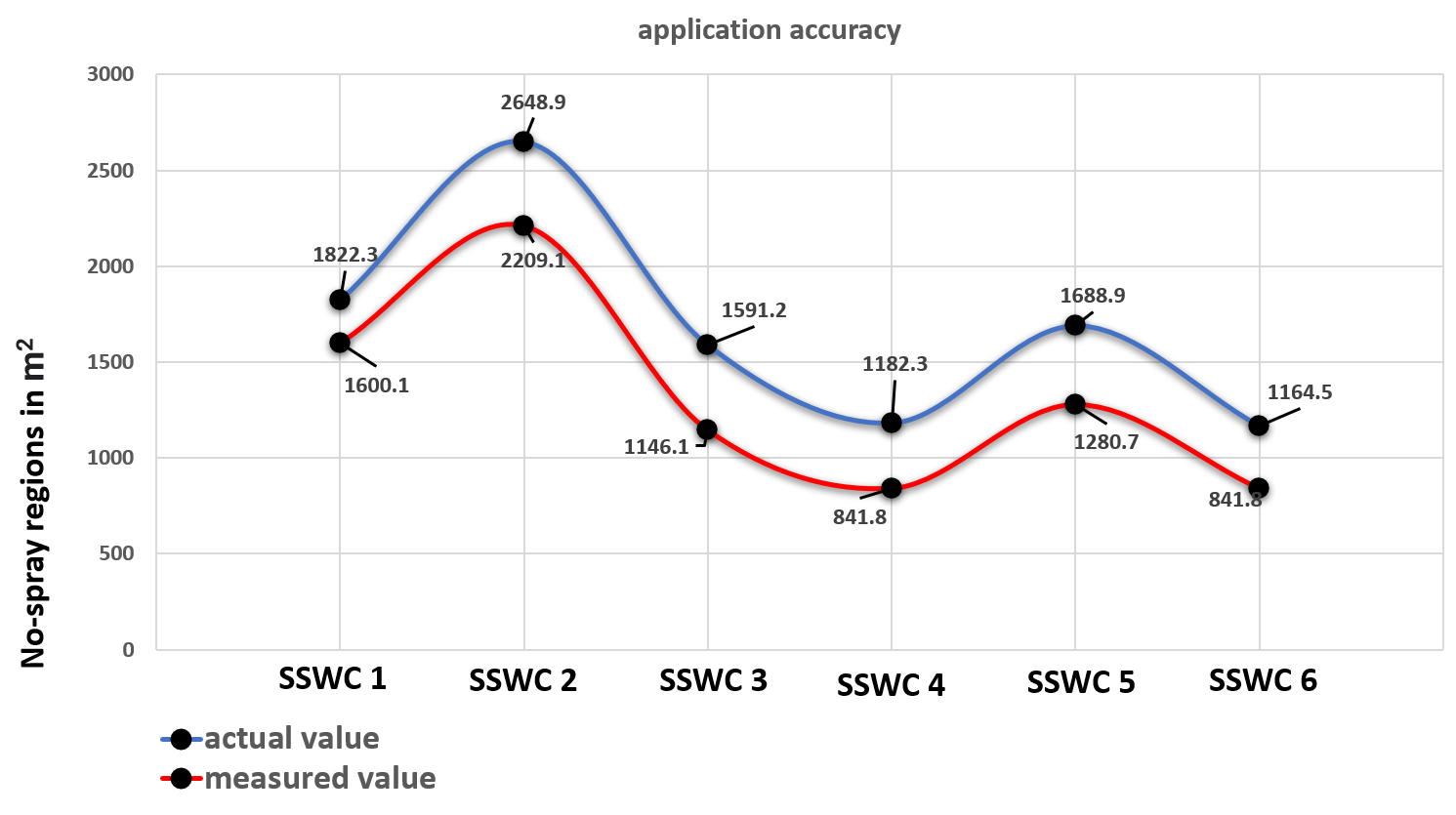} }}%
    \qquad
    \caption{Graphical representation of actual value and measured value of land area that was supposed to be saved from being sprayed with chemical herbicide in the 6 SSWC treatment plots  }%
    \label{figure8}
    \end{figure}

Overall, our study was able to achieve an application accuracy of 78.42\% in terms of not-spraying in the regions where there was no presence of weeds. Figure \ref{figure8} shows the no-spray land distribution information that was designed and implemented for 6 SSWC treatment plots. During the field implementation of this approach through a commercial-size sprayer, the overlap between individual nozzles could be a reason why the application accuracy is not much higher. Although the spray nozzle footprint covers the space between the nozzles (0.5 m), it is likely that the application rate towards the middle of the spraying pattern is higher than on its edges.

\begin{figure}[h!]
    \centering
    \includegraphics[width=0.9\linewidth]{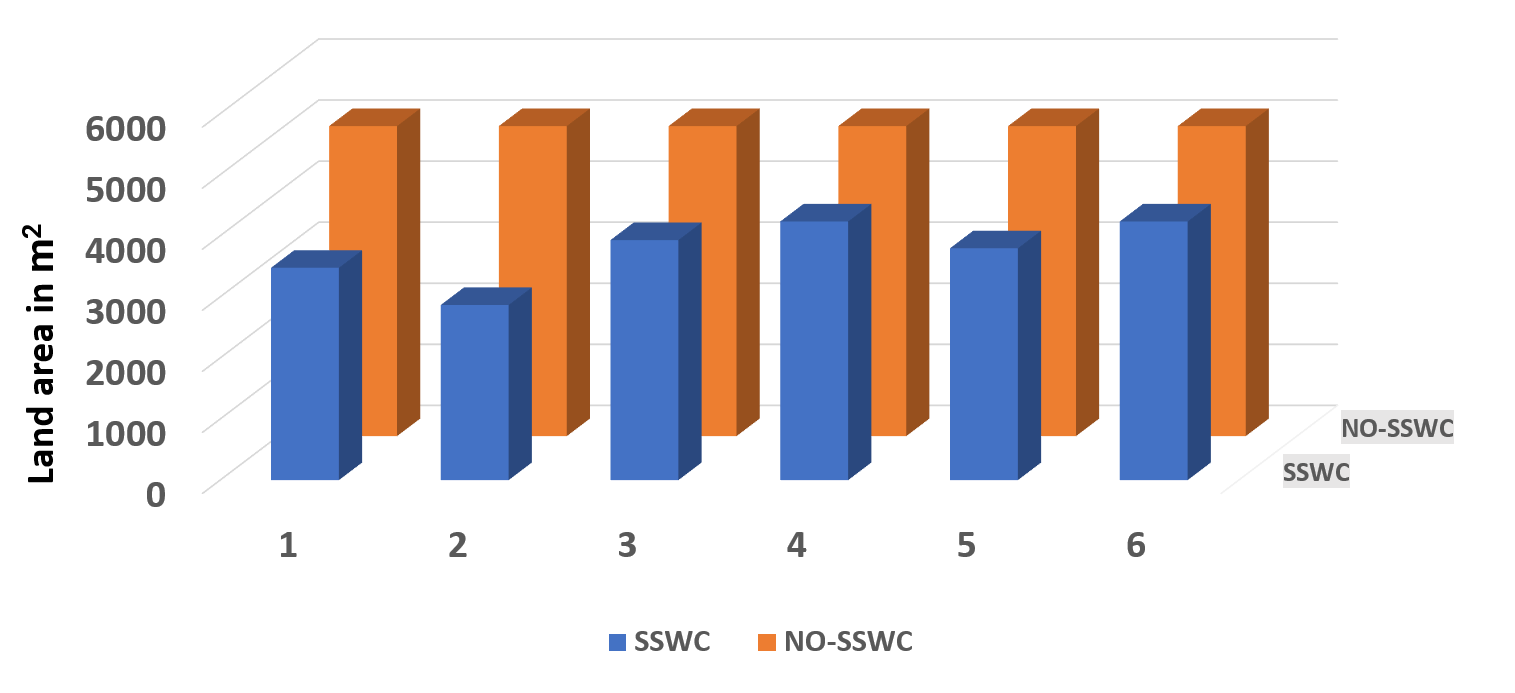}
    \caption{Bar graph showing land area that was sprayed with chemical herbicide in SSWC and no-SSWC (conventional) treatment plots}
    \label{figure9}
\end{figure}

Likewise, our approach for SSWC achieved significant chemical savings. Using the prescription map to turn off the sprayer’s nozzles on the grids cells free of weed resulted in savings of 26\% of the area to be sprayed. Our method saved 7919.6 $m^{2}$ out of 30457 $m^{2}$ land from being sprayed with the chemical herbicide. Figure \ref{figure9} shows the consumption of chemicals on the basis of land area using our SSWC approach and the conventional or no-SSWC approach. Although the weed pressure at the season we performed this study was unusually high, the study proved that it is possible to spray chemicals only to the necessary part of the field. Since the conventional method that applies chemicals uniformly across the field would spray chemicals in the 7919.6  $m^{2}$ land inside SSWC while our approach did not spray any chemical in that area, therefore our study was able to save around 26\% of chemical herbicide in the six SSWC treatment plots compared to the existing conventional method. Our approach minimized the chemical usage to this value in a practical agricultural environment of corn production system in agriculture.

Because of the significant reduction of chemical usage, our approach of spraying chemical herbicide only to those area where there is presence of weed could be a helpful practice in making this environmental ecosystem less disturbed from being attacked with agro-chemicals. Since 75\% of all planted acres of cropland across the US are row crops, the approach we adopted in this study could be a foundational study for performing SSWC in the vast majority of US row crops. On the other hand, our approach made a financial savings on that 26\% reduced chemical. Moreover, the reduction of land area for spraying herbicides directly reduces the amount of water usage, therefore, our approach also made significant saving(approximately 26\%) in the water usage for chemical spray application. 

\par Although some leading agricultural machinery manufacturing companies such as John Deere and CNH Industrials have announced about their future release of an automated sprayer which could perform site-specific application in some crops including corn, the technology upon successful release would be adopted by few farmers only because it is not that easy for all farmers to switch into newer technology due to the expensive cost of agricultural spray machines. However, farmers could easily use our approach with the existing variable rate sprayers in the market that could read the prescription maps.  

In a post harvest imagery based field evaluation of the SSWC and no-SSWC treatments plots, we found that the treatment regions were significantly different in terms of weed presence $\left(\mathrm{F}_{1,5}=11.41, \mathrm{p}<0.0197\right)$. The amount of weeds present in SSWC treatment plots in terms of surface area was 3.4 times higher than the amount of weeds that were present in no-SSWC plots. The findings from this study could be a foundation to pursue various research questions that our study could not address, because of time and other limiting factors. Although the imagery obtained from the field had high resolution (GSD= 0.63 cm per pixel), we might have missed smaller weeds than the camera resolution during the data collection of the early growing season. Those weed detection skips might be one of the reasons why we observed weeds in the SSWC plots after the corn harvest. One could argue that image resolution should be increased to avoid those skips, but there are other challenges to achieve that, such as cost of better hardware (UASs and sensors), time to collect data (by flying lower with the same equipment used on this study), and time to process the data (lower altitude flight would lead to many more pictures captured, which increases processing time). The reasons to why the SSWC had more weeds than no-SSWC could likely be a combination of factors: 1) weeds were sprayed too late (rate not adequate to the size of the weeds); 2) lack of overlap when individual nozzles were applying chemical (applying even less chemical to the already too big/tall weeds); and 3) missing small weeds that germinated after we mapped the area.

\section{Conclusion and future suggestions}
According to the result achieved in this study regarding chemical savings, our approach of fusing UAS data with spraying platform have the potential to minimize chemical consumption in weed control application in the agricultural production system of the US corn. Additionally, our SSWC approach have the potential to reduce financial expenditure that farmers are currently investing in chemicals to apply conventional approach or no-SSWC approach in corn production. On the other hand, production of corn using our approach appear to be more healthy than the corn being produced through conventional weed control approach because everyone would love to consume corn that is being produced by using less chemicals. 

The approach proposed in this study regarding identifying corn rows and removing them from the UAS imagery showed to be an effective way to map the weeds during the early growing season of corn. However, this study was performed through human intervention for weed mapping which would take around a day to collect and process data for making a weed prescription map. Although we have beaten the state-of-the art and achieved some strong milestones in the investigations that have been performed for site-specific weed management implementation using UAS platform, to make our SSWC approach consumer grade in commercial agricultural production system, our future suggestions are:
\begin{itemize}
    \item Automating real time prescription map generation from the information collected using UAS could reduce the data collection and processing time. For that, use of embedded platforms to integrate UAS data with sprayer platform could be a field to explore in the near future.
    \item The overlap between individual nozzles of the sprayer should be set in a way that the spray-pattern covers the entire region set in the prescription map. 
    \item Since the sprayer was operated at lower speed(11.5 km/hr approximate) than the normal practice(19 km/hr approximate), in commercial farming, approximate speed of because of the GPS refreshing capability of the the sprayer. Upon improvement of the sprayer's GPS refreshing rate, the application could be improved to operate at higher speed.
\end{itemize}

\section{Acknowledgements and Funding}

The authors would like to acknowledge Richard Richter and Chad Richter for operating the commercial sprayer during field implementation of this study. The project was funded by North Dakota Corn Council (NDCC). 

\section*{Author contributions statement}
Investigation, conceptualization, formal analysis, data analysis, writing and editing: Ranjan Sapkota, Paulo Flores;  Data analysis, reviewing: John Stenger, Michael Ostlie

\bibliographystyle{unsrtnat}
\bibliography{references}  






\end{document}